\documentclass[dvipsnames,format=sigconf,anonymous=false,review=false,nonacm]{acmart}

\AtBeginDocument{%
  \providecommand\BibTeX{{%
    \normalfont B\kern-0.5em{\scshape i\kern-0.25em b}\kern-0.8em\TeX}}}

\copyrightyear{2023}
\acmYear{2023}
\setcopyright{rightsretained}
\acmConference[GECCO '23]{Genetic and Evolutionary Computation Conference}{July 15--19, 2023}{Lisbon, Portugal}
\acmBooktitle{Genetic and Evolutionary Computation Conference (GECCO '23), July 15--19, 2023, Lisbon, Portugal}
\acmDOI{10.1145/3583131.3590380}
\acmISBN{979-8-4007-0119-1/23/07}

\usepackage[utf8]{inputenc} 
\usepackage[T1]{fontenc}    
\usepackage{hyperref}       
\usepackage{url}            
\usepackage{booktabs}       
\usepackage{amsfonts}       
\usepackage{nicefrac}       
\usepackage{microtype}      

\usepackage[ruled,vlined,linesnumbered]{algorithm2e} 
\usepackage{xcolor}
\def\HiLi{\leavevmode\rlap{\hbox to \hsize{\color{gray!25}\leaders\hrule height .8\baselineskip depth .5ex\hfill}}}

\usepackage{comment}
\usepackage{mathtools}
\usepackage{cuted}
\usepackage{bm}
\usepackage{bbm}
\usepackage{pifont}
\usepackage{dsfont}
\usepackage{comment}
\usepackage{caption}
\usepackage{threeparttable}
\usepackage{siunitx}
\usepackage{tikz}
\usetikzlibrary{shapes.geometric, arrows, fit}
\usepackage{subcaption}

\ifdefined\N                                                                
\renewcommand{\N}{\mathds{N}}                                                
\else
  \newcommand{\N}{\mathds{N}}
\fi
\newcommand{\R}{\mathds{R}}                                                 
\ifdefined\C
  \renewcommand{\C}{\mathds{C}}                                             
\else
  \newcommand{\C}{\mathds{C}}
\fi

\newcommand{\xv}{\mathbf{x}}													

\renewcommand{\P}{\mathds{P}}                                               
\newcommand{\E}{\mathds{E}}                                                 

\newcommand{\Xspace}{\mathcal{X}}                                           
\newcommand{\Yspace}{\mathcal{Y}}                                           
\newcommand{\Pxy}{\P_{xy}}                                                  
\newcommand{\allDatasets}{\mathds{D}}                                       
\newcommand{\D}{\mathcal{D}}                                                      
\newcommand{\defAllDatasets}{\bigcup_{n \in \N}(\Xspace \times \Yspace)^n}  
\renewcommand{\xi}[1][i]{\mathbf{x}^{(#1)}}                                          
\newcommand{\yi}[1][i]{y^{(#1)}}                                            

\newcommand{\preimageInducerShort}{\allDatasets\times\bm{\Lambda}}     
\newcommand{\inducer}{\mathcal{I}}                                                

\newcommand{\Hspace}{\mathcal{H}}														
\newcommand{\fh}{\hat{f}}                                                   
\newcommand{\fhDlambda}{\fh_{\D, \lambdav}}                                       
\newcommand{\lambdav}{\bm{\lambda}}											

\newcommand{\Ilam}{\inducer_{\lambdav}}						

\newcommand{\repo}{\url{https://github.com/slds-lmu/paper_2023_eagga}}

\begin{document}

\title{Multi-Objective Optimization of Performance and Interpretability of Tabular Supervised Machine Learning Models}


\author{Lennart Schneider}
\affiliation{%
  \institution{LMU Munich \& Munich Center for Machine Learning (MCML)}
  \streetaddress{Ludwigstraße 33}
  \city{Munich}
  \country{Germany}
  \postcode{80539}
}
\email{lennart.schneider@stat.uni-muenchen.de}

\author{Bernd Bischl}
\affiliation{%
  \institution{LMU Munich \& Munich Center for Machine Learning (MCML)}
  \streetaddress{Ludwigstraße 33}
  \city{Munich}
  \country{Germany}
  \postcode{80539}
}
\email{bernd.bischl@stat.uni-muenchen.de}

\author{Janek Thomas}
\affiliation{%
  \institution{LMU Munich \& Munich Center for Machine Learning (MCML)}
  \streetaddress{Ludwigstraße 33}
  \city{Munich}
  \country{Germany}
  \postcode{80539}
}
\email{janek.thomas@stat.uni-muenchen.de}


\begin{abstract}
We present a model-agnostic framework for jointly optimizing the predictive performance and interpretability of supervised machine learning models for tabular data.
Interpretability is quantified via three measures: feature sparsity, interaction sparsity of features, and sparsity of non-monotone feature effects.
By treating hyperparameter optimization of a machine learning algorithm as a multi-objective optimization problem, our framework allows for generating diverse models that trade off high performance and ease of interpretability in a single optimization run.
Efficient optimization is achieved via augmentation of the search space of the learning algorithm by incorporating feature selection, interaction and monotonicity constraints into the hyperparameter search space.
We demonstrate that the optimization problem effectively translates to finding the Pareto optimal set of groups of selected features that are allowed to interact in a model, along with finding their optimal monotonicity constraints and optimal hyperparameters of the learning algorithm itself.
We then introduce a novel evolutionary algorithm that can operate efficiently on this augmented search space.
In benchmark experiments, we show that our framework is capable of finding diverse models that are highly competitive or outperform state-of-the-art XGBoost or Explainable Boosting Machine models, both with respect to performance and interpretability.
\end{abstract}



\maketitle

\section{Introduction}\label{sec:intro}
Tabular data are highly relevant for numerous application areas such as finance, bio-informatics, and medical diagnosis.
State-of-the-art learning algorithms for tabular data include tree-based methods, e.g., gradient boosted trees (with larger depth) \citep{friedman2001greedy} such as XGBoost \citep{xgboost} and LightGBM \citep{lightgbm}, or random forests \citep{Breiman_2001}, which often still outperform deep neural networks \citep{Grinsztajn_Oyallon_Varoquaux_2022}, although the performance gap has recently shrunk considerably \citep{Grinsztajn_Oyallon_Varoquaux_2022,Gorishniy_Rubachev_Khrulkov_Babenko_2021,Kadra_Lindauer_Hutter_Grabocka_2021,Shwartz-Ziv_Armon_2022}.
To achieve peak predictive performance, AutoML tools such as AutoGluon-Tabular \citep{Erickson_Mueller_Shirkov_Zhang_Larroy_Li_Smola_2020} or AutoSklearn \citep{feurer2015efficient} often make further use of ensembling and stacking multiple models.
Moreover, careful hyperparameter optimization of learning algorithms is typically required to yield well performing models \citep{Probst_Boulesteix_Bischl_2019,Rijn_Hutter_2018}.

While good predictive performance is generally of central importance, many applications desire or even require models to fulfill additional criteria, such as \emph{interpretability} or \emph{sparseness}.
For example a model used for medical diagnosis that achieves high accuracy but lacks interpretability, such as black box models like gradient boosted trees or deep neural networks, may encounter difficulties in gaining trust and adoption.
In contrast, a model that can provide insights into its reasoning, even if it has slightly lower performance, is more likely to be trusted and used in real-world scenarios.
In the field of Interpretable Machine Learning \citep{molnar2022}, two different approaches for achieving \emph{interpretability} of models have broadly emerged: \textbf{(i)} to only consider learning algorithms that induce ``interpretable'' models due to their simple intrinsic nature (e.g., logistic regression, decision trees, rule-based systems or generalized additive models) or \textbf{(ii)} to use post-hoc methods  -- which can either be model-agnostic, such as partial dependence plots (PDP) \citep{friedman2001greedy} or accumulated local effects (ALE) \citep{apley2020visualizing}, or model-specific -- to gain insight into the inner workings of a model.

When working with tabular data in real-world situations, finding the ``right'' model can be cumbersome and involves time-consuming manual trial and error.
Often, various learning algorithms are tried to produce different models, which are then inspected to select a final model based on concrete user preferences at hand.
While this process may be feasible if the goal is to ``simply'' find a good-performing model, it becomes inefficient if additional criteria such as feature \emph{sparseness}, few \emph{interactions} of features, or \emph{monotonicity} of feature effects are also to be considered.
In particular, monotonicity can be highly relevant in practice, as frequently only a model consistent with domain knowledge is acceptable to domain experts.
For example, in credit loan approval, models are often required to be monotone with respect to the decision variables involved \citep{Velikova2004DecisionTF}.
Our framework allows automatic generation of a set of models that balance performance and \emph{interpretability}.
Formally, this requires two things: \textbf{(i)} a way to measure the interpretability of models on a global scale, and \textbf{(ii)} an efficient approach for solving the arising \emph{multi-objective} optimization problem.

\textbf{Our Contributions.} We introduce a general, model-agnostic framework for jointly optimizing the predictive performance and interpretability of supervised machine learning models for tabular data.
To achieve this, we propose a quantification of the \emph{interpretability} of models on a global scale based on three measures: feature sparsity, interaction sparsity of features, and sparsity of non-monotone feature effects.
We then formulate a multi-objective optimization problem of performance and interpretability over the hyperparameter search space of a learning algorithm, which is augmented by incorporating feature selection as well as interaction and monotonicity constraints into the hyperparameter search space.
As a solution to the optimization problem, we present a novel hyperparameter optimization algorithm that can operate efficiently on this augmented search space, making use of the principles of evolutionary computation by treating feature selection as well as the specification of interaction and monotonicity constraints of features as a grouping problem.

\section{Related Work}\label{sec:literature}

When choosing a learning algorithm that induces interpretable models -- e.g., logistic regression models, Elastic-Nets \citep{Zou_Hastie_2005}, or generalized additive models (GAMs) \citep{hastie1986} -- one typically loses predictive performance compared to black box models obtained via, e.g., tree based ensembles \citep{Couronne_Probst_Boulesteix_2018}.
However, the downside of these black box models is that their interpretability is hindered by potentially plenty of interaction effects of features and non-linear or non-monotone feature effects.
The Explainable Boosting Machine (EBM) \citep{lou2012intelligible,Lou_2013} positions itself between comparably poor-performing but intelligible models and well-performing but unintelligible models.
EBM is a tree-based, cyclic gradient boosting GAM using automatic interaction detection based on FAST \citep{Lou_2013} to include a given number of second-order interactions in the model.
EBM often yields good predictive performance \citep{Nori_Jenkins_Koch_Caruana_2019} while being more intelligible than black box models.
Nevertheless, EBM has some drawbacks: \textbf{(i)} EBM is comparably slow to train, as it relies on a large number of boosting steps with a small learning rate to cycle through all features\footnote{Which we also observed in our benchmark experiments.}, \textbf{(ii)} EBM naturally cannot induce a sparse model, as all features are included in a round robin fashion, and the contribution of each feature to a final prediction is therefore non-zero, \textbf{(iii)} as a result of the large number of boosting steps, EBM often fits highly non-linear and non-monotone shape functions (resulting in rather complex relationships of features and target), and, relatedly, \textbf{(iv)} EBM cannot handle monotonicity constraints during training -- i.e., if it is known (or even required) that a feature should have a monotone increasing effect on the target variable, EBM can neither make use of this information nor guarantee such an effect.

A popular approach for constructing sparser models is given by feature selection, which is also related to the complexity and intelligibility of a model \citep{Guyon_Elisseeff_2003,bischl2010selecting,Binder_Moosbauer_Thomas_Bischl_2020}.
While feature selection can also be performed in the context of unsupervised learning \citep{handl2006feature}, we focus on the supervised learning context.
Here, the goal of feature selection is to select only a subset of relevant features while still constructing a model with good predictive performance.
There are two model-agnostic approaches to feature selection \citep{Guyon_Elisseeff_2003}: feature filters and feature wrappers.
Feature filters use proxy measures that are cheap to compute to rank features by their potential explanatory power independent of the concrete learning algorithm being used.
Popular examples include measures based on information theory, correlation, distance, or consistency \citep{Dash_Liu_1997}.
In contrast to feature filters, feature wrappers directly optimize predictive performance over the space of feature subsets \citep{Kohavi_John_1997}.
As every feature subset evaluation requires one or multiple model fits, making exhaustive search infeasible, a discrete black box optimization search strategy (such as a greedy search or an evolutionary algorithm \citep{Xue_Zhang_Browne_Yao_2016}) is necessary.
On the one hand, feature selection is often considered a single-objective optimization problem, and the feature selection step is only used to optimize performance \citep{Kohavi_John_1997}.
On the other hand, feature selection can also be framed as a multi-objective optimization problem, maximizing predictive performance and feature sparsity simultaneously \citep{xue2014multi,Binder_Moosbauer_Thomas_Bischl_2020}.
Finally, recent work also explored the idea of identifying sets of features without predefined grouping \citep{imrie2022compfs}.

Looking at measures for interpretability of models on a global scale, Molnar and colleagues \citep{Molnar_Casalicchio_Bischl_2020} were among the first to explicitly propose model-agnostic measures of model complexity.
They quantify model complexity by decomposing the prediction function of any model into a sum of components with increasing dimensionality, based on which they derive three measures: the number of features used by a model, the interaction strength of features, and the main effect complexity of features.


\section{Theoretical Background}\label{sec:background}
Consider the supervised learning problem of inferring a model from labeled data $\D$ with $n$ observations where each observation $(\xi, \yi)$ consists of a $p$-dimensional feature vector $\xi$.
We assume that $\D$ has been sampled i.i.d.\ from an underlying, unknown distribution, $\D \sim (\Pxy)^n$.
A learning algorithm or \emph{inducer} $\inducer$ configured by hyperparameters $\lambdav \in \bm{\Lambda}$ maps a data set $\D$ to a model $\fh$, i.e., $\inducer : \preimageInducerShort \to \Hspace, (\D, \lambdav) \mapsto \fhDlambda$, where $\allDatasets := \defAllDatasets$ is the set of all data sets, $\bm{\Lambda}$ is the search space of hyperparameters, and $\Hspace$ is the hypothesis space of models.
In general, one is interested in constructing a model $\fhDlambda = \inducer(\D, \lambdav)$ that minimizes the \emph{generalization error}\footnote{With a slight abuse of notation, we will write $\Ilam$ to denote that a certain hyperparameter configuration $\lambdav$ is fixed, i.e., $\Ilam(\D) = \inducer(\D, \lambdav)$ with $\lambdav$ fixed.}, $\mathrm{GE}(\fhDlambda) = \E_{(\xv, y) \sim \Pxy} \left[L(\fhDlambda(\xv), y)]\right]$, where $L$ is a loss function measuring discrepancy between the prediction and true label.
However, the generalization error can only be estimated using in-sample data, $\widehat{\mathrm{GE}}(\Ilam, \D)$, through a resampling technique such as cross-validation.
For more details, see, e.g., \cite{feurer2019hyperparameter,Bischl_2021}.

\subsection{Multi-Objective Hyperparameter Optimization}
Let $c_{1}: \bm{\Lambda} \to \R, \ldots, c_{m}: \bm{\Lambda} \to \R, m \in \N$ denote $m$ evaluation criteria of machine learning models.
Note that evaluation criteria usually also depend on the data set and resampling technique at hand (which we omit here for clarity).
Define $c: \bm{\Lambda} \to \R^m$ to assign an $m$-dimensional cost vector to a hyperparameter configuration $\lambdav \in \bm{\Lambda}$. 
The general multi-objective hyperparameter optimization problem is then defined as $\min_{\lambdav \in \bm{\Lambda}} c(\lambdav) =  \min_{\lambdav \in \bm{\Lambda}} \left( c_1(\lambdav), c_2(\lambdav), \ldots, c_m(\lambdav) \right)$.
Generally, there is no single hyperparameter configuration that minimizes all criteria, as these criteria typically compete with one another.
Therefore, focus is given to the concept of Pareto optimality and the set of Pareto optimal configurations:
A hyperparameter configuration $\lambdav \in \bm{\Lambda}$ \emph{(Pareto-)dominates} another configuration $\lambdav^\prime \in \bm{\Lambda}$, written as $\lambdav  \prec \lambdav^\prime$, if and only if
\begin{align*}
  \begin{split}
  \forall i \in \left\{1, \ldots m\right\}&: c_i\left(\lambdav\right) \le c_i\left(\lambdav^\prime\right)\, \land \\
  \exists j \in \left\{1, \ldots m\right\}&: c_j\left(\lambdav\right) < c_j\left(\lambdav^\prime\right).
  \end{split}
\end{align*}
The set of Pareto optimal solutions is therefore defined as $\mathcal{P} := \left\{\lambdav \in \bm{\Lambda} ~|~ \not \exists~ \lambdav^\prime \in \bm{\Lambda} \text{ s.t. } \lambdav^\prime \prec \lambdav\right\}$.
The image of $\mathcal{P}$ under $c$, $c(\mathcal{P})$, is called the Pareto front.
The goal of multi-objective optimization is to find a set of configurations $\mathcal{\hat{P}}$ so that $c(\mathcal{\hat{P}})$ approximates the true Pareto front well.

A popular quality indicator of multi-objective optimization is given by the dominated Hypervolume \citep{zitzler1998multiobjective}.
The Hypervolume of an approximation of the Pareto front $c(\mathcal{\hat{P}})$ is defined as the combined volume of the dominated hypercubes of all solution points with respect to a reference point $\bm{r} \in \R^m$.
For more details on multi-objective hyperparameter optimization in general as well as an overview of recent applications, we refer to \cite{karl2022multi,morales2022survey}.

\subsection{Quantifying Interpretability}\label{sec:measures}
We propose a quantification of interpretability that is conceptually similar to \citep{Molnar_Casalicchio_Bischl_2020}, but our measures and their operationalization differ.
As measures for the interpretability of a model on a global scale, we propose to use feature sparsity, interaction sparsity of features, and sparsity of non-monotone features.
All our measures are based on the prediction function $\hat{f}: \Xspace_{} \rightarrow \R^g$ of a model\footnote{For regression, $g$ is $1$, while in classification the output usually represents the $g$ decision scores or posterior probabilities of the $g$ candidate classes. Without loss of generalization, we will assume $g = 1$ in the following.}.

To define whether feature $j$ is used by the model, we can determine whether the prediction function changes if the value of $x_{j}$ changes, i.e.,
$\hat{f}(x_{1}, \ldots, x_{j}^{\prime}, \ldots x_{p}) \neq \hat{f}(x_{1}, \ldots, x_{j}, \ldots x_{p})$ whenever $x_{j}^{\prime} \neq x_{j}$.
The (relative) number of features used by a model, $NF$, can then be defined as
\begin{equation}
\begin{split}
NF(\hat{f}) \coloneqq  | \{&j \in \{1, \ldots, p\}: \exists x_{j}, x_{j}^{\prime} \in \Xspace_{j}, x_{j}^{\prime} \neq x_{j}~\text{s.t.}~\\
& \hat{f}(x_{1}, \ldots, x_{j}^{\prime}, \ldots x_{p}) \neq \hat{f}(x_{1}, \ldots, x_{j}, \ldots x_{p})\}| / p.
\end{split}
\end{equation}

Similarly, we want to define whether two features $j$ and $k$ interact.
A prediction function $\hat{f}$ of a model exhibits an interaction between two features $j$ and $k$ if the difference in the value of $\hat{f}(\mathbf{x})$ as a result of changing the value of $x_{j}$ depends on the concrete value of $x_{k}$ \citep{friedman2008predictive}.
Consequently, given no interaction of features $j$ and $k$, $\hat{f}$ can be decomposed into $\hat{f}(\mathbf{x}) = f_{-j}(\mathbf{x}_{-j}) + f_{-k}(\mathbf{x}_{-k})$ where $\mathbf{x}_{-j}$ and $\mathbf{x}_{-k}$ are feature vectors excluding $x_{j}$ and respectively $x_{k}$.
The (relative) number of interactions in a model, $NI$, can then be defined as
\begin{equation}
\begin{split}
NI(\hat{f}) \coloneqq | \{&\{j, k\}, j, k \in \{1, \ldots, p\}, k > j: \nexists f_{-j}, f_{-k}~\text{s.t.}~\\
& \hat{f}(\mathbf{x}) = f_{-j}(\mathbf{x}_{-j}) + f_{-k}(\mathbf{x}_{-k})\}| / ((p (p - 1))/2).
\end{split}
\end{equation}

If the hypothesis space of an inducer is restricted to only contain models including main effects and second-order interaction effects of features, $NI$ is a direct measure of the violation of interaction sparsity of a model.
However, if the hypothesis space contains models that include higher order interaction effects, $NI$ falls short in penalizing such higher order interactions.
To penalize the inclusion of many pairwise interactions and higher order interactions, we assume transitivity with respect to the interaction of features, i.e., if feature $j$ and $k$ and $k$ and $l$ interact, we also count an interaction of feature $j$ and $l$.

Finally, we define feature $j$ to have a monotone increasing effect if it holds that whenever $x_{j} \le x_{j}^{\prime}$, one has that $\hat{f}(x_{1}, \ldots, x_{j}, \ldots x_{p}) \le \hat{f}(x_{1}, \ldots, x_{j}^{\prime}, \ldots x_{p})$.
Analogously, we define feature $j$ to have a monotone decreasing effect.
The (relative) number of non-monotone features in a model, $NNM$, is then given by
\begin{equation}
\begin{split}
NNM(\hat{f}) \coloneqq | \{&j \in \{1, \ldots, p\}: (\exists x_{j}, x_{j}^{\prime} \in \Xspace_{j}, x_{j} \le x_{j}^{\prime}~\text{s.t.}~\\
& \hat{f}(x_{1}, \ldots, x_{j}, \ldots x_{p}) > \hat{f}(x_{1}, \ldots, x_{j}^{\prime}, \ldots x_{p}))~\land \\
& (\exists x_{j}, x_{j}^{\prime} \in \Xspace_{j}, x_{j} \le x_{j}^{\prime}~\text{s.t.}~\\
& \hat{f}(x_{1}, \ldots, x_{j}, \ldots x_{p}) < \hat{f}(x_{1}, \ldots, x_{j}^{\prime}, \ldots x_{p}))\}| / p.
\end{split}
\end{equation}

Based on these formal definitions, $NF$, $NI$, and $NNM$ can be \emph{operationalized} in different ways.
For example, $NF$ can be estimated via a sampling procedure, as described in \cite{Molnar_Casalicchio_Bischl_2020}.
Similarly, $NI$ could in principle be estimated based on the partial dependence function \citep{friedman2008predictive} or by calculating H-statistics \citep{friedman2008predictive} or Greenwell's interaction index \citep{greenwell2018simple} for all pairs of features.
%
%
%
Depending on the concrete learning algorithm at hand, $NF$ and $NI$ can often also be determined in a straightforward manner by, e.g., looking at features used in splits in a decision tree.
In the following, we will exactly determine $NF$ and $NI$ by directly inspecting the resulting model whenever possible.
Finally, looking at monotonicity, estimating $NNM$ is arguably difficult.
In principle, one could try to test whether a feature has a monotone effect via verification-based testing \citep{sharma2020testing} or adaptive random testing \citep{chen2005}.
However, such procedures are always at risk of error, and as monotonicity is typically a hard\footnote{In practice, a feature is typically expected to exhibit a monotone effect, or not, without any in-between or probabilistic formulation.} requirement of a model \citep{potharst2002classification,Velikova2004DecisionTF}, we opt to determine $NNM$ based on the configuration of the inducer.
This requires the inducer to allow for the specification of monotonicity constraints of features, which is easily achievable for, e.g., tree-based methods or GAMs.

We want to note that a model that has low values with respect to $NF$, $NI$ and $NNM$ still can be complex and must not necessarily result in being intrinsically interpretable.
Nevertheless, we believe that such a model is much more easier to interpret, e.g., based on a post-hoc ALE analysis, compared to a model with high values in $NF$, $NI$, or $NNM$.
For instance, if a model uses only few features that have monotone increasing effects and do not interact with each other, the prediction function of the model can be easily summarized.
For example, increasing the value of any individual feature would result in an increase in the predicted outcome, regardless of the values of other features.
Such a simple and consistent relationship between features and the predicted outcome makes the model more \emph{interpretable}.
This direct connection between model \emph{complexity} and ease of interpretability is also the reason why we deem it appropriate to speak of multi-objective optimization of performance and \emph{interpretability}.


\subsection{Multi-Objective Optimization of Performance and Interpretability}\label{sec:problem}

We formulate the hyperparameter optimization problem of a learning algorithm as a multi-objective optimization problem with the goal of minimizing the estimated generalization error, $NF$, $NI$ and $NNM$.
To allow for efficient optimization, we extend the search space of the learning algorithm and include hyperparameters for the selection of features, interaction constraints, and monotonicity constraints of features to be part of the search space.
Therefore, we require the learning algorithm to allow for the specification of feature selection as well as interaction and monotonicity constraints of features.

In the following, we denote by $\bm{\check{\Lambda}}$ the extended search space.
A hyperparameter configuration $\bm{\check{\lambda}} \in \bm{\check{\Lambda}}$ is given by the tuple $(\bm{\lambda}, \bm{s}, \bm{I}_{\bm{s}}, \bm{m}_{\bm{I}_{\bm{s}}})$.
Here, $\bm{\lambda} \in \bm{\Lambda}$ is the usual hyperparameter configuration of a learning algorithm, $\bm{s}$ is a binary vector of length $p$, indicating selection of features, $\bm{I}_{\bm{s}}$ is a symmetric matrix of dimension $p \times p$ with $(\bm{I}_{\bm{s}})_{jk} = 1$ indicating that features $j$ and $k$ are allowed to interact in a model and $0$ indicating otherwise, and $\bm{m}_{\bm{I}_{\bm{s}}}$ is an integer vector of length $p$ indicating monotonicity constraints of features ($-1$ for monotone decreasing, $1$ for monotone increasing, and $0$ for unconstrained\footnote{We will later argue that it suffices to only consider $\{0, 1\}$ as monotonicity constraints.}).

In principle, we could proceed to try solving the multi-objective optimization problem as given in Equation~\ref{eq:mo_orig}:
\begin{equation}\label{eq:mo_orig}
    \min_{\bm{\check{\lambda}} \in \bm{\check{\Lambda}}} \left( \widehat{\mathrm{GE}}\left(\inducer_{\bm{\check{\lambda}}}, \D\right), NF\left(\hat{f}_{\D, \bm{\check{\lambda}}}\right), NI\left(\hat{f}_{\D,\bm{\check{\lambda}}}\right), NNM\left(\hat{f}_{\D, \bm{\check{\lambda}}}\right) \right)
\end{equation}
Although this formulation of the optimization problem is quite natural, it has several drawbacks:
First, note that the extended search space has become complex, including a binary vector, a quadratic matrix, and an integer vector that scale linearly or quadratic in the number of features $p$. 
Second, note that $\bm{I}_{\bm{s}}$ depends on $\bm{s}$, as only features that have been selected can be allowed to interact.
Similarly, $\bm{m}_{\bm{I}_{\bm{s}}}$ depends on both $\bm{I}$ and $\bm{s}$.
For example, if feature $j$ is required to have a monotone increasing effect but is also allowed to interact with another feature $k$, then the monotonicity of feature $j$ may not be guaranteed if feature $k$ does not also have a monotone increasing effect.
This is because the interaction between feature $j$ and $k$ can potentially alter the overall effect of feature $j$, and without the monotonicity constraint on feature $k$, the monotonicity of feature $j$ may be compromised.
Therefore, in the general model-agnostic case, it is most straightforward to require both features $j$ and $k$ to have monotone increasing effects to ensure that the monotonicity of feature $j$ is maintained in the presence of their potential interaction effect.

We will now derive a reformulation of the search space of the optimization problem stated in Equation~\ref{eq:mo_orig} that is much easier to handle.
To do so, recall the definition of an endorelation and the properties reflexive, symmetric, and transitive.
Note that a reflexive, symmetric, and transitive endorelation -- also called an equivalence relation -- imposes a group structure on a set, i.e., it partitions the set by means of its equivalence classes.

To arrive at an easier formulation of the search space of the optimization problem in Equation~\ref{eq:mo_orig}, we define interactions of features as an endorelation.
Let $C = \{1, \ldots, p\}$ denote the index set of features and $C_{s} \subseteq C$ the index set of features selected for inclusion in a model and define an endorelation $R$ on $C_{s}$, $R \subseteq C_{s} \times C_{s}$.
We say feature $j$ and feature $k$ are \emph{allowed to interact} if the model in principle allows for the inclusion of an (interaction) effect of the two, and write $jRk$.
It follows that $R$ is naturally reflexive and symmetric -- i.e., if feature $j$ is allowed to interact with feature $k$, then the reverse also holds, as the interaction of features is non-directional.
However, note that the interaction of features must in fact not be transitive -- i.e., even if feature $j$ and $k$ and $k$ and $l$ interact in a model, it must not follow that feature $j$ and $l$ also interact.
Nevertheless, from a modeling perspective, it is reasonable to \emph{allow} for features $j$ and $l$ to also interact, partially also due to the potential presence of a three-way interaction, which (in the most general scenario) can only be included in a model if $R$ is closed under transitivity (and the same argument can be made for higher-order interactions)\footnote{This is also directly related to the principle of marginality; see, e.g., \cite{Nelder1977}.}.
It is therefore natural to always consider the transitive closure of $R$, resulting in an equivalence relation.
This implies that the equivalence classes induced by $R$ partition the index set of selected features and naturally call for working with a \emph{group structure}.
Regarding monotonicity constraints of features, we want to note that monotonicity constraints must simply be defined as attributes of the equivalence classes (for the same reason illustrated earlier: if features are allowed to interact, they should share the same monotonicity constraint).

We can now introduce the group structure space $\bm{\mathcal{G}}$.
Each group structure $\bm{G} \in \bm{\mathcal{G}}$ consists of a $g$-tuple of sets of feature indices with the first set, i.e., group, representing the features that were not selected ($C \setminus C_{s}$) and all remaining sets resembling the $k$ equivalence classes under the equivalence relation $R \subseteq C_{s} \times C_{s}$ of features being \emph{allowed to interact} with each equivalence class also being equipped with a monotonicity attribute.
Any group structure can therefore be encoded as follows: $\bm{G} = (G_{1} = C \setminus C_{s}, G_{2} = (E_{1}, M_{E_{1}}), \ldots, G_{g} = (E_{k}, M_{E_{k}}))$.
Here, $E_{k} \subseteq C_{s}$ is an index set containing the indices of features part of the $k$-th equivalence class under $R$, and $M_{E_{k}} \in \{-1, 0, 1\}$ is the monotonicity attribute of the $k$-th equivalence class.
We can now reformulate Equation~\ref{eq:mo_orig} and introduce the augmented search space $\bm{\tilde{\Lambda}} = \bm{\Lambda} \times \bm{\mathcal{G}}$ by considering the group structure $\bm{G} \in \bm{\mathcal{G}}$ instead of $\bm{s}$, $\bm{I}_{\bm{s}}$,  and $\bm{m}_{\bm{I}_{\bm{s}}}$.
The reformulated search space now consists of the Cartesian product of the search space of the learning algorithm, $\bm{\Lambda}$, and the group structure space $\bm{\mathcal{G}}$ and each configuration, $\bm{\tilde{\lambda}}$ of the search space is given by a tuple $(\bm{\lambda}, \bm{G})$, which we argue is much easier to optimize.
We visualize the components involved in the optimization problem in Figure~\ref{fig:api}.
\definecolor{lightblue}{RGB}{173,216,229}
\tikzstyle{inducer} = [rectangle, rounded corners, minimum width=2cm, minimum height=1cm,text centered, draw=black, fill=gray!30]
\tikzstyle{data} = [rectangle, rounded corners, minimum width=1cm, minimum height=1cm,text centered, draw=black, fill=white]
\tikzstyle{hyperparams} = [rectangle, rounded corners, minimum width=1cm, minimum height=1cm,text centered, draw=black, fill=white]
\tikzstyle{model} = [rectangle, rounded corners, minimum width=1cm, minimum height=1cm,text centered, draw=black, fill=white]
\tikzstyle{constr} = [rectangle, rounded corners, minimum width=2cm, minimum height=1cm,text centered, draw=black, fill=lightblue]
\tikzstyle{arrow} = [thick,->,>=stealth]
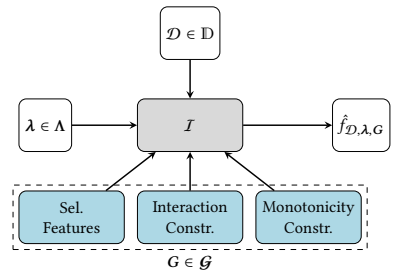
\begin{figure}[ht]
\centering
\scalebox{0.7}{
\begin{tikzpicture}[node distance=1.75cm]
  \node (data) [data] {$\D \in \allDatasets$};
  \node (inducer) [inducer, below of=data] {$\inducer$};
  \node (hyperparams) [hyperparams, left of=inducer, xshift=-1cm] {$\lambdav \in \bm{\Lambda}$};
  \node (model) [model, right of=inducer, xshift=1.5cm] {$\hat{f}_{\D, \lambdav, \bm{G}}$};
  \node (interactions) [constr, align=center, below of=inducer] {Interaction\\Constr.};
  \node (features) [constr, align=center, left of=interactions, xshift=-0.5cm] {Sel.\\Features};
  \node (monotone) [constr, align=center, right of=interactions, xshift=0.5cm] {Monotonicity\\Constr.};
  \node [draw=black, label=below:$\bm{G} \in \bm{\mathcal{G}}$, fit={(features) (interactions) (monotone)}, dashed] {};
  \draw [arrow] (data) -- (inducer);
  \draw [arrow] (hyperparams) -- (inducer);
  \draw [arrow] (features) -- (inducer);
  \draw [arrow] (interactions) -- (inducer);
  \draw [arrow] (monotone) -- (inducer);
  \draw [arrow] (inducer) -- (model);
\end{tikzpicture}
}
\caption{Overview of the components involved in the hyperparameter optimization problem. The inducer is required to allow for the specification of feature selection, as well as interaction and monotonicity constraints of features, which are derived based on the group structure $\bm{G} \in \bm{\mathcal{G}}$.}
\label{fig:api}
\end{figure}

\section{Method}\label{sec:method}

For optimizing the multi-objective optimization problem, we introduce an optimizer consisting of an evolutionary algorithm (EA) for the original search space of the learning algorithm $\bm{\Lambda}$ and a so-called grouping genetic algorithm (GGA) \citep{Falkenauer_1993} for the group structure space $\bm{\mathcal{G}}$.
We therefore dub our optimizer \emph{EAGGA}.

\subsection{EAGGA}\label{sec:eagga}
The combination of using an EA and GGA allows us to jointly operate on the augmented search space $\bm{\tilde{\Lambda}} = \bm{\Lambda} \times \bm{\mathcal{G}}$.
\emph{EAGGA}'s main routine is heavily inspired by NSGA-II \citep{Deb_2002}. 
NSGA-II is an evolutionary multi-objective algorithm making use of the concepts of non-dominated sorting and crowding distance to select individuals for survival close to the Pareto front that also cover a wide spread along the Pareto front.
In each generation, NSGA-II iterates through reproduction, crossover, mutation, and survival steps that generate the population of the next generation.
In \emph{EAGGA}, we perform parent selection via a binary tournament selection and simply apply suitable crossover and mutation operators to hyperparameters of the original search space ($\bm{\lambda} \in \bm{\Lambda}$) and group structures ($\bm{G} \in \bm{\mathcal{G}}$) next to each other to produce offspring.


\subsubsection{EA Operators}\label{sec:ga_operators}
For the original hyperparameters of the learning algorithm ($\lambdav \in \bm{\Lambda}$), we use the Cartesian product of operators that operate in different ways on the different parameter types \citep{Li_2013}.
We use a global crossover probability of $p = 0.7$ and a global mutation probability of $p = 0.3$.
All hyperparameters undergo uniform crossover ($p = 0.5$) for recombination.
Numeric and integer hyperparameters undergo Gaussian mutation ($p = 0.2, \sigma = 0.1$; values min-max scaled to $[0, 1]$ prior to mutation and re-transformed afterwards; values rounded to the closest integer in the case of integer hyperparameters), while categorical hyperparameters undergo uniform mutation ($p = 0.2$).
The choice of operators and probabilities of crossover and mutation were mostly inspired by \citep{Binder_Moosbauer_Thomas_Bischl_2020}.

\subsubsection{GGA Operators}\label{sec:gga_operators}
Group structures ($\bm{G} \in \bm{\mathcal{G}}$) undergo mutation and crossover operators inspired by the original work of Falkenauer \citep{Falkenauer_1993,Falkenauer_1996}.
We again use a global crossover probability of $p = 0.7$ and a global mutation probability of $p = 0.3$.
Recall that a group structure is encoded as $\bm{G} = (G_{1} = C \setminus C_{s}, G_{2} = (E_{1}, M_{E_{1}}), \ldots, G_{g} = (E_{k}, M_{E_{k}}))$ where $G_{1} = C \setminus C_{s}$ is an index set of features not selected and each $E_{k} \subseteq C_{s}$ is an index set of features part of the $k$-th equivalence class under the equivalence relation $R$ of features being \emph{allowed to interact}, and $M_{E_{k}} \in \{-1, 0, 1\}$ is the monotonicity attribute of the $k$-th equivalence class.
The basic idea of a GGA is to apply operators directly on the group structure.
For crossover, we select two crossing sites, delimiting the crossing section, in each of the two parents (e.g., $G_{1} G_{2} | G_{3} | G_{4}$ and $H_{1} | H_{2} H_{3} | H_{4} H_{5}$; $G$ used for the first parent and $H$ for the second parent).
We then inject the contents (groups together with their monotonicity attributes) of the crossing section of the first parent at the first crossing site of the second parent (e.g., inserting  $G_{3}$ into the second parent, resulting in $H_{1} G_{3} H_{2} H_{3} H_{4} H_{5}$). 
Finally, we remove all items (feature indices) from the old groups now occurring twice in the second parent.
For example, assume $H_{3} = (\{1, 2, 3\}, 0)$ and $G_{3} = (\{3\}, 1)$, then after inserting $G_{3}$ into the second parent, $H_{3}$ is given by $(\{1, 2\}, 0)$.
In the case of the first group, i.e., the index set of features not selected, being injected, we simply add these indices to the first group of the parent.
To create the second offspring, we swap the roles of the parents.
For more details on the GGA crossover, see \cite{Falkenauer_1996}.
For mutation, we simply assign each feature index a new group membership with probability $p = 0.2$ and sample a new monotonicity attribute for each group with probability $p = 0.2$.
To allow for more precise handling of the group structure, we incorporate a feedback loop into \emph{EAGGA}:
After evaluating an offspring, we can determine the actual features and interactions (closed under transitivity) as included in the model\footnote{The group structure only imposes an upper constraint, meaning that the resulting model may use all or some of the selected features, and the same applies to interactions.} and update the group structure $\bm{G}$ of each offspring.
In Section~\ref{sec:ablation} and the supplementary material, we present results of an ablation study investigating the effect of turning off either crossover or mutation of group structures or both, where we observed that in general both of them are needed for good performance.

\subsection{Initializing the Group Structures}\label{sec:init}
As hyperparameter optimization is costly, we strive to make \emph{EAGGA} more sample-efficient.
We use three detectors (feature, interaction, and monotonicity) to find better initial population group structures.
An ablation study in Section~\ref{sec:ablation} shows that these detectors substantially improve \emph{EAGGA}'s (anytime) performance.

\subsubsection{Feature Detector}\label{sec:init_feature}
The goal of a \emph{feature detector} is to quantify the importance of features so that the probability of selecting an important feature $j$ (i.e., $j \in C_{s}$) can be increased.
Formally, a feature detector maps a data set $\D$ to a $p$-dimensional vector of real valued scores with the $j$-th element corresponding to the score of the $j$-th feature.
In \emph{EAGGA}, we use feature filters. 
A feature filter measures feature importance using a fast proxy, such as the entropy-based information gain filter \citep{Largeron_2011}, which calculates the difference between the target variable's entropy and the joint entropy conditioned on the feature.
Based on the filter score for each feature, we can then weight the probability of selecting a feature.
To determine the number of selected features $S$ of a member of the initial population, we sample a random integer between $1$ and $p$ from a truncated geometric distribution similarly as in \citep{Binder_Moosbauer_Thomas_Bischl_2020}.
The features that are actually selected are then determined by sampling from all binary vectors $\bm{s}$ of length $p$ that sum to $S$ with weighted probabilities according to the feature filter scores.

\subsubsection{Interaction Detector}\label{sec:init_inter}
The idea of a (pairwise) \emph{interaction detector} is to quantify the importance of interactions of features so that the probability of those features being in the same group (i.e., the same equivalence class under the equivalence relation $R$ \emph{allowed to interact}) can be increased.
Formally, an interaction detector maps a data set $\D$ to a symmetric, real valued $p \times p$ matrix with the element at the $j$-th row and $k$-th column corresponding to the score of the $j$-th and $k$-th feature\footnote{Note that the diagonal is of no interest and can be set to, e.g., $0$.}.
Recall that in \emph{EAGGA}, the first group $G_{1}$ of a group structure $\bm{G}$ is always given by the indices of features that are not selected.
To initialize the remaining groups, we make use of the FAST algorithm \citep{Lou_2013}.
FAST allows for efficient quantification of the importance of all pairwise interactions of features based on the residual sums of squares when extending a main effects model to include an interaction effect.
To determine the number of included interactions $I$ of a member of the initial population, we sample a random integer between $1$ and $(p (1 - p)) /2$ from a truncated geometric distribution.
The actual groups are then determined by considering the $I$ most important pairwise interactions according to FAST, constructing an equivalence relation $R$ \emph{allowed to interact}, and deriving the equivalence classes under $R$.

\subsubsection{Monotonicity Detector}\label{sec:init_mon}
Using a \emph{monotonicity detector} is helpful due to two reasons:
First, recall that the monotonicity attribute of a group can in principle either be -1 (monotone decreasing), 1 (monotone increasing), or 0 (unconstrained).
This is somewhat redundant, as a monotone decreasing feature effect (without loss of generalization, we assume purely numeric features) can always be realized by enforcing a monotone increasing effect and swapping the sign of the feature itself.
Therefore, by detecting whether a monotone feature effect should be increasing or decreasing we can encode monotonicity constraints more efficiently.
Second, by quantifying the mismatch in model fit between enforcing monotonicity and no constraint, the monotonicity detector can bias the probability of the monotonicity attribute being unconstrained.
Formally, a monotonicity detector maps a data set $\D$ to a $p$-dimensional vector of real valued scores with the $j$-th element corresponding to the score of the $j$-th feature where the sign of the score indicates the direction of monotonicity and the magnitude of the score reflects the strength of the monotone relationship between the feature and the target variable.
In \emph{EAGGA}, we use the following monotonicity detector:
For each feature, we fit a decision tree on sub-sampled data and obtain the predictions.
We then calculate Spearman's $\rho$ between the feature values and the target predictions.
Finally, we repeat this process 10 times and calculate the average Spearman's $\rho$, which we scale\footnote{This is done to allow for some non-determinism during sampling.} to $[0.2, 0.8]$.
For each group of features of a member of the initial population, we take the average over the individual scores and use this average as a probability to sample the monotonicity attribute of the group.

\section{Benchmark Experiments}\label{sec:benchmarks}
To our best knowledge, \emph{EAGGA} is the first model-agnostic approach to perform \emph{efficient} multi-objective optimization of performance and interpretability of machine learning models by incorporating feature selection as well as interaction and monotonicity constraints into the hyperparameter search space.
In our experiments, we combine \emph{EAGGA} with XGBoost (\emph{EAGGA}\textsubscript{XGBoost}) or XGBoost with a maximum depth fixed to 2 (\emph{EAGGA}\textsubscript{XGBoost\textsubscript{md2}}, resulting in second-order interactions being the most complex higher-order interactions that can be picked up by the model).
We configure \emph{EAGGA} to use a population size of $\mu = 100$ and an offspring size of $\nu = 10$, with the comparably large population size being inspired by \citep{xue2014multi,Binder_Moosbauer_Thomas_Bischl_2020}.
One naïve approach to generate a benchmark baseline is to simply use a collection of competitors that all excel at different objectives which \emph{EAGGA} tries to optimize jointly and compare \emph{EAGGA}\textsubscript{XGBoost} to the union of the competitors.
Another approach is to compare \emph{EAGGA}\textsubscript{XGBoost} to standard multi-objective optimization of XGBoost (without augmentation of the search space).
Code and supplementary material are released via \repo.

\subsection{\emph{EAGGA} vs. A Collection of Competitors}\label{sec:benchmarks_competitors}
We construct a collection of competitors by considering an EBM, Elastic-Net, (untuned) random forest, and XGBoost.
An EBM offers good performance with few interactions, an Elastic-Net provides sparse, monotone solutions, while a random forest and XGBoost usually deliver strong results using many features, interactions, and non-monotone effects.
We tune the hyperparameters of the EBM, Elastic-Net, and XGBoost via Bayesian Optimization\footnote{We employ a Bayesian Optimization variant similarly configured as SMAC \citep{lindauer2022smac3}, i.e., using a random forest as surrogate model and Expected Improvement \citep{Jones_Schonlau_Welch_1998} as acquisition function.} and optimize for predictive performance.
For the search spaces of the learning algorithms, see our supplementary material.
All learning algorithms are given a budget of 8 hours of sequential runtime on a single CPU (note that this is a disadvantage for \emph{EAGGA}, as each competitor is given the same computational budget and therefore the union of competitors uses substantially more compute budget than \emph{EAGGA}).
As a performance metric, we choose the area under the receiver operating characteristic curve (AUC)\footnote{We minimize the negative AUC.}.
Performance estimation is conducted via nested resampling: As an outer resampling, we use a holdout with a ratio of $2/3$, i.e., test performance is evaluated on $1/3$ of the data.
Hyperparameter optimization is then performed using 5-fold cross-validation on the remaining $2/3$ of the data.
For \emph{EAGGA}\textsubscript{XGBoost} and \emph{EAGGA}\textsubscript{XGBoost\textsubscript{md2}}, the Pareto optimal configurations found during optimization are re-evaluated on the test-set.
For the EBM, Elastic-Net, random forest, and XGBoost, we re-evaluate the single best-performing configuration (found during optimization) on the test-set.
For XGBoost models, $NF$ and $NI$ are determined by actually checking the model and all splits in all trees, whereas $NNM$ is determined based on the monotonicity constraints of features used in the model (only applicable when optimized via \emph{EAGGA}; for the standard XGBoost, we assume $NNM$ to be the same as $NF$ as we consider monotonicity of features to be a hard requirement as explained in Section~\ref{sec:measures}). 
For the EBM, $NF$ is always $1$, as EBM cycles through all available features in a round robin fashion, whereas $NI$ is directly given by the value of the hyperparameter \texttt{interactions} and we assume $NNM$ to be the same as $NF$, as EBM does not allow for the specification of monotonicity constraints and cannot guarantee monotone feature effects.
For the Elastic-Net, $NF$ is determined by looking at the relative number of non-zero coefficients, whereas $NI$ and $NNM$ are always $0$ (no interaction effects are included in the standard Elastic-Net and feature effects are always monotone).
Finally, for the random forest, $NF$ and $NI$ are again determined by actually checking the model and all splits in all trees, whereas $NNM$ is again the same as $NF$ (for the same reason as for the standard XGBoost).

All methods are compared on twenty binary classification tasks taken from OpenML CC-18 \citep{Bischl_Openml_2021} and the AutoML benchmark \citep{Gijsbers_Bueno_Coors_LeDell_Poirier_Thomas_Bischl_Vanschoren_2022}.
We perform 10 replications of each optimization run on each task with different random seeds to allow for statistical analysis.
Criteria for selecting the tasks were fewer than $100000$ observations, the number of features being fewer than $1000$ as well as numeric features, i.e., we focus on small- to medium-sized tabular data sets.
We only consider binary classification tasks, as the EBM until now does not support the inclusion of interaction effects of features in the case of multi-class classification.
More details on the data sets can be found in our supplementary material.

\begin{figure}[ht]
\centering
\includegraphics[width=\linewidth]{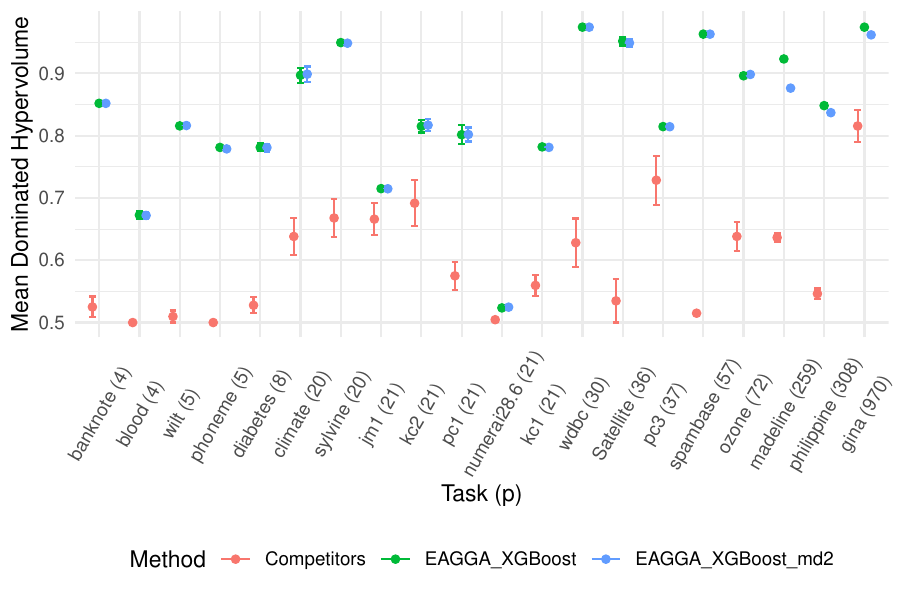}
\caption{Mean dominated Hypervolume of \emph{EAGGA}\textsubscript{XGBoost}, \emph{EAGGA}\textsubscript{XGBoost\textsubscript{md2}}, and the union of competitors averaged over 10 replications. Bars represent standard errors.}
\label{fig:dhv}
\end{figure}

As we are comparing a multi-objective optimization framework (\emph{EAGGA}) to a collection of models, we perform the following analysis:
For every run on each task, we calculate the dominated Hypervolume of the (test-set) Pareto front of \emph{EAGGA}\textsubscript{XGBoost} and \emph{EAGGA}\textsubscript{XGBoost\textsubscript{md2}} with respect to the reference point $\bm{r} = (0, 1, 1, 1)^\top$ and compare this with the dominated Hypervolume obtained by considering the non-dominated set of the EBM, Elastic-Net, random forest, and XGBoost solutions (evaluated on the test-set).
To allow for a fair comparison, we always include a featureless learner that simply predicts the majority class without relying on any features when calculating the dominated Hypervolume\footnote{As the resulting point $(- 0.5, 0, 0, 0)^\top$ will have a large contribution to the dominated Hypervolume, but only \emph{EAGGA} might be able to consistently find a hyperparameter configuration resulting in such a model.}.
Results are given in Figure~\ref{fig:dhv}.
Note that the number in parentheses after a task name indicates the number of features of the task.
Using \emph{EAGGA} results in substantially larger dominated Hypervolume (Wilcoxon signed-ranks test \citep{demvsar2006statistical} on the mean dominated Hypervolume over replications: $T = 0, p < 0.001$ for \emph{EAGGA}\textsubscript{XGBoost} vs. competitors and $T = 0, p < 0.001$ for \emph{EAGGA}\textsubscript{XGBoost\textsubscript{md2}} vs. competitors).

We further determine for each task the fraction of replications where each competitor yields a solution that is Pareto-dominated by the solutions of \emph{EAGGA}\textsubscript{XGBoost} or \emph{EAGGA}\textsubscript{XGBoost\textsubscript{md2}}.
Table~\ref{tab:dominated_a} shows this fraction averaged over all tasks for \emph{EAGGA}\textsubscript{XGBoost} -- i.e., on average, roughly $46\%$ of the EBM solutions are Pareto-dominated by the solutions found by \emph{EAGGA}\textsubscript{XGBoost}.
Table~\ref{tab:dominated_b} shows this fraction averaged over all tasks for \emph{EAGGA}\textsubscript{XGBoost\textsubscript{md2}}.
We also compute the counterpart -- i.e., what is the fraction of replications where the whole Pareto set of \emph{EAGGA}\textsubscript{XGBoost} or \emph{EAGGA}\textsubscript{XGBoost\textsubscript{md2}} is dominated by the Pareto set of the union of the competitors.
This was never the case, neither for \emph{EAGGA}\textsubscript{XGBoost} nor \emph{EAGGA}\textsubscript{XGBoost\textsubscript{md2}}.
We want to note that in some runs, evaluating the initial design during optimization of the EBM took longer than the whole compute budget of 8 hours.
In these cases, our fallback was to only evaluate the default configuration suggested by the EBM authors.

In our supplementary material, we also provide an illustrative example of the usage of \emph{EAGGA} relying on the \emph{ozone-level-8hr} task and analyze an exemplary Pareto front.
Additionally we analyze the best performing models from each method in terms of AUC and interpretability.
Results show that the best models found by \emph{EAGGA} perform similarly to XGBoost models optimized for performance, but use less features, interactions, and non-monotone features, indicating improved interpretability.

\begin{table}[ht]
  \centering
  \caption{Mean fraction of runs over tasks and replications where competitors yield a solution that is dominated by \emph{EAGGA}\textsubscript{XGBoost} or \emph{EAGGA}\textsubscript{XGBoost\textsubscript{md2}}.}
  \label{tab:dominated}
  \footnotesize
  \begin{subtable}[t]{.5\linewidth}
  \centering
  \caption{\emph{EAGGA}\textsubscript{XGBoost}}
  \label{tab:dominated_a}
  \begin{threeparttable}
  \begin{tabular}{l r r}
    \toprule
    \textbf{Competitor} & \textbf{Mean} & \textbf{SE}\\
    \midrule
    EBM & 0.46 & 0.04\\
    Elastic-Net & 0.30 & 0.03\\
    Random Forest & 0.81 & 0.03\\
    XGBoost & 0.40 & 0.03\\
    \bottomrule
  \end{tabular}
  \begin{tablenotes}
    \tiny
    \item SE = standard error.
  \end{tablenotes}
  \end{threeparttable}
  \end{subtable}%
  \begin{subtable}[t]{.5\linewidth}
  \centering
  \caption{\emph{EAGGA}\textsubscript{XGBoost\textsubscript{md2}}}
  \label{tab:dominated_b}
  \begin{threeparttable}
  \begin{tabular}{l r r}
    \toprule
    \textbf{Competitor} & \textbf{Mean} & \textbf{SE}\\
    \midrule
    EBM & 0.36 & 0.03\\
    Elastic-Net & 0.28 & 0.03\\
    Random Forest & 0.74 & 0.03\\
    XGBoost & 0.31 & 0.03\\
    \bottomrule
  \end{tabular}
  \begin{tablenotes}
    \tiny
    \item SE = standard error.
  \end{tablenotes}
  \end{threeparttable}
  \end{subtable}%
\end{table}

\subsection{\emph{EAGGA} vs. Multi-Objective XGBoost}\label{sec:benchmarks_xgboost_mo}

\begin{figure}[ht]
\centering
\includegraphics[width=\linewidth]{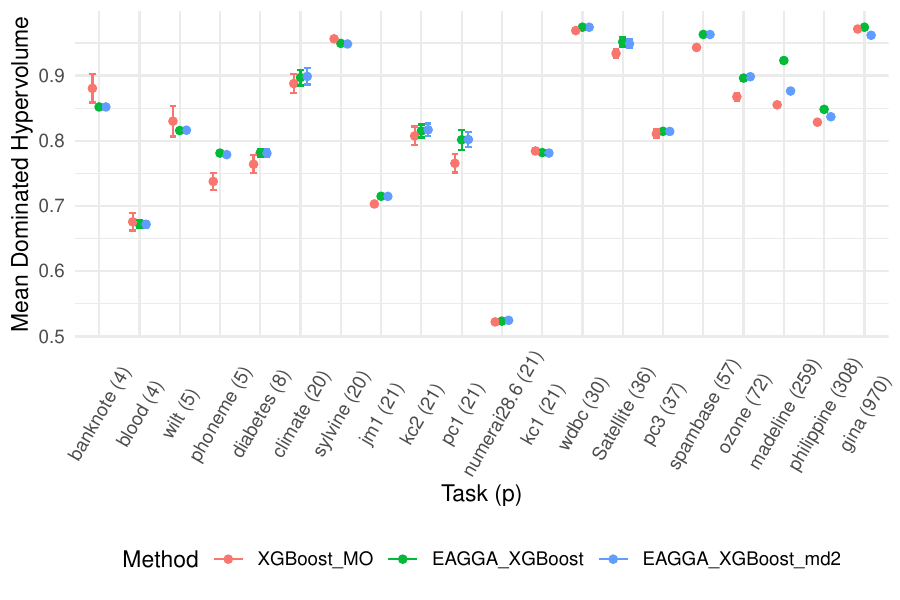}
\caption{Mean dominated Hypervolume of \emph{EAGGA}\textsubscript{XGBoost}, \emph{EAGGA}\textsubscript{XGBoost\textsubscript{md2}}, and XGBoost\textsubscript{MO} averaged over 10 replications. Bars represent standard errors.}
\label{fig:dhv_xgboost}
\end{figure}

We also compare \emph{EAGGA}\textsubscript{XGBoost} to multi-objective optimization of XGBoost (without augmentation of the search space), which we will refer to as XGBoost\textsubscript{MO}.
As an optimizer, we employ ParEGO \citep{Knowles_2006}, a scalarization-based multi-objective Bayesian Optimization algorithm that we configure to use a random forest as surrogate model and Expected Improvement as acquisition function.
The search space used within ParEGO is exactly the same as the search space used within \emph{EAGGA} -- with the exception that we do not augment the search space to include feature selection, interaction, and monotonicity constraints, as standard multi-objective optimizers such as ParEGO cannot naturally operate on such a search space.
The question we want to answer is whether it is sufficient to work on the standard search space with a standard multi-objective optimizer to optimize XGBoost for predictive performance and interpretability.
Benchmark tasks and the evaluation protocol are exactly the same as in Section~\ref{sec:benchmarks_competitors} -- i.e., for \emph{EAGGA}\textsubscript{XGBoost}, \emph{EAGGA}\textsubscript{XGBoost\textsubscript{md2}}, and XGBoost\textsubscript{MO}, the Pareto optimal configurations found during optimization are re-evaluated on the test-set.
For each run on each task, we calculate the dominated Hypervolume of the (test-set) Pareto front of \emph{EAGGA}\textsubscript{XGBoost}, \emph{EAGGA}\textsubscript{XGBoost\textsubscript{md2}}, and XGBoost\textsubscript{MO}, which we visualize in Figure~\ref{fig:dhv_xgboost}.
Again, using \emph{EAGGA} results in usually at least the same and often substantially larger dominated Hypervolume (Wilcoxon signed-ranks test on the mean dominated Hypervolume over replications: $T = 40, p = 0.0076$ for \emph{EAGGA}\textsubscript{XGBoost} vs. XGBoost\textsubscript{MO} and $T = 50, p = 0.02$ for \emph{EAGGA}\textsubscript{XGBoost\textsubscript{md2}} vs. XGBoost\textsubscript{MO}).
Notably, the only tasks where XGBoost\textsubscript{MO} outperforms \emph{EAGGA} are tasks with few features.
In our supplementary material, we also analyze the anytime dominated Hypervolume during optimization (i.e., calculated on the inner resampling).

\subsection{An Ablation Study of \emph{EAGGA}}\label{sec:ablation}

We perform an ablation study of the components of \emph{EAGGA} with the goal to answer the following questions:
\textbf{(i)} Does \emph{EAGGA} improve over a random search on the same search space?
\textbf{(ii)} How important are crossover and respectively mutation of group structures?
\textbf{(iii)} What is the benefit of using detectors to initialize the population?

To do so, we rerun all benchmark experiments with different flavors of \emph{EAGGA} and analyze the mean dominated Hypervolume during optimization, i.e., calculated on the inner resampling.
We consider the following modifications or ``flavors'' of \emph{EAGGA}:
\textbf{(i)} Simply performing a random search on $\tilde{\bm{\Lambda}}$ after using \emph{EAGGA}'s detectors to initialize the population (\texttt{Random Search}).
\textbf{(ii)} Switching off either crossover or mutation of group structures ($\bm{G} \in \bm{\mathcal{G}}$) or both (\texttt{No\_Crossover}, \texttt{No\_Mutation}, \texttt{No\_Cross\_Mut}).
\textbf{(iii)} Switching off the detectors of \emph{EAGGA} and initializing the population at random (\texttt{No\_Detectors}).

\begin{figure}[ht]
\centering
\includegraphics[width=\linewidth]{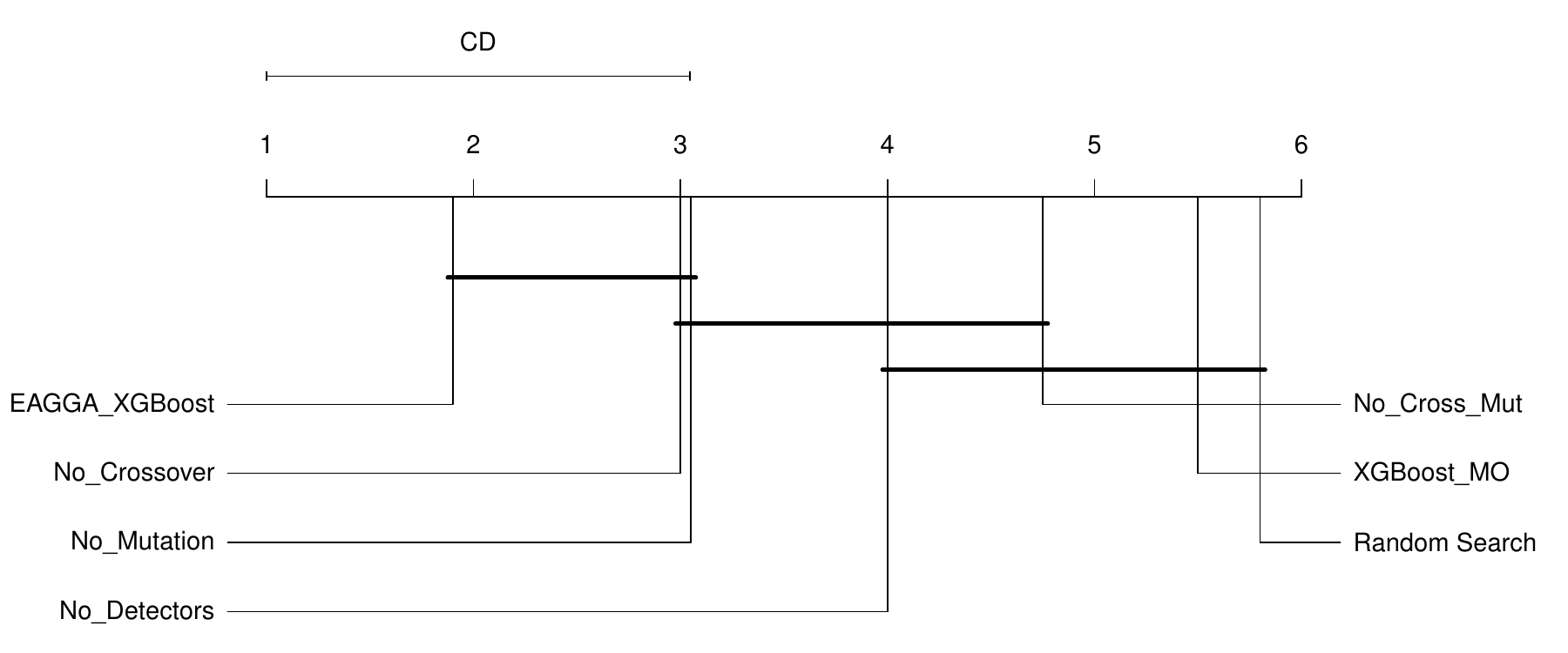}
\caption{Critical difference plot of the ranks of optimizers based on the final mean dominated Hypervolume during optimization. Lower rank is better.}
\label{fig:eagga_ablation_cd_1}
\end{figure}

We observe that \textbf{(i)} performing a random search performs comparably poorly, \textbf{(ii)} crossover and mutation of group structures are needed for good performance and \textbf{(iii)} using detectors can boost the performance although this is mainly due to using detectors strongly affecting the early performance of \emph{EAGGA}.
Conducting a Friedman test \citep{demvsar2006statistical} on the final mean dominated Hypervolume during optimization indicates significant differences in ranks of optimizers ($\chi^{2}(6) = 52.99, p < 0.001$).
Figure~\ref{fig:eagga_ablation_cd_1} visualizes the corresponding critical difference plot based on the follow up Nemenyi test.
For completeness, we also include XGBoost\textsubscript{MO}.
For detailed results and discussion, please see our supplementary material.

\section{Conclusion}\label{sec:conclusion}
We have presented a general model-agnostic framework for jointly optimizing the predictive performance and interpretability of supervised machine learning models for tabular data.
\emph{EAGGA} is a multi-objective optimizer making use of the principles of evolutionary computation to jointly optimize the hyperparameters of a learning algorithm as well as the group structure of features.
\emph{EAGGA} allows for obtaining a set of diverse models in a single optimization run and can outperform state-of-the-art competitors both with respect to performance and interpretability.

In practice, users may have prior knowledge about which features to include, which features should interact or even a requirement for a certain feature to have a monotone effect.
Although we studied \emph{EAGGA} in the context of no prior knowledge, it can be extended to incorporate such information by initializing the population accordingly and preventing crossover and mutation from creating offspring incongruent with the prior.


\emph{EAGGA} might be especially useful when using deep neural networks as learning algorithms, as Kadra and colleagues \cite{Kadra_Lindauer_Hutter_Grabocka_2021} demonstrated that strong regularization of neural networks can be a key component to achieving good performance on tabular data.
Using \emph{EAGGA} in combination with neural networks would require the design of a network architecture that allows for the specification of interaction and monotonicity constraints of features.
Notable work in this direction has been undertaken by \citep{tsang2018neural,yang2021gami,chang2022node,radenovic2022neural,dubey2022scalable}.

Finally, it must be noted that \emph{EAGGA} cannot guarantee that the resulting group structure of a model is sensible, and the structure must be verified by domain experts (with respect to the selection of features, as well as their interaction and monotonicity constraints).
Nevertheless, we believe that \emph{EAGGA} can be of significant interest for a wide variety of users.

\begin{acks}
The authors of this work take full responsibilities for its content.
Lennart Schneider is supported by the Bavarian Ministry of Economic Affairs, Regional Development and Energy through the Center for Analytics - Data - Applications (ADACenter) within the framework of BAYERN DIGITAL II (20-3410-2-9-8).
Lennart Schneider acknowledges funding from the LMU Mentoring Program of the Faculty of Mathematics, Informatics and Statistics.
\end{acks}

\bibliographystyle{ACM-Reference-Format}
\bibliography{references}

\appendix

\section{Illustrative Example}\label{sec:appendix_example}
We illustrate the potential of our approach using a concrete example, relying on the \emph{ozone-level-8hr} task.
The goal is to predict whether a day is a high ozone day or not using 72 features such as temperature measured at different time throughout the day.
We again use \emph{EAGGA} with XGBoost and compare it to an EBM, Elastic-Net, random forest, and XGBoost.
The evaluation protocol is the same as used in the benchmark experiments.
We summarize the results in Table~\ref{tab:ozone}.
\emph{EAGGA} is able to find a good spread of models that trade off performance and interpretability to varying degree.
For example, the best-performing XGBoost model found by \emph{EAGGA} is close to the performance of an XGBoost model solely optimized for performance but uses substantially fewer features, interactions, and non-monotone features.

\begin{table}[ht]
  \centering
  \caption{Pareto front obtained using \emph{EAGGA\textsubscript{XGBoost}} on the \emph{ozone-level-8hr} task compared to the solutions found using an EBM, Elastic-Net, random forest, or XGBoost.}
  \label{tab:ozone}
  \footnotesize
  \begin{subtable}[t]{.5\linewidth}
  \centering
  \caption{\emph{EAGGA}\textsubscript{XGBoost}}
  \label{tab:ozone_eagga}
  \begin{threeparttable}
  \begin{tabular}{l r r r}
    \toprule
    \textbf{AUC} & $\mathbf{NF}$  & $\mathbf{NI}$ & $\mathbf{NNM}$ \\
    \midrule
    0.802 & 0.014 & 0.000 & 0.000\\
    0.818 & 0.083 & 0.002 & 0.000\\
    0.829 & 0.139 & 0.005 & 0.000\\
    0.831 & 0.042 & 0.000 & 0.028\\
    0.841 & 0.042 & 0.000 & 0.042\\
    0.863 & 0.097 & 0.008 & 0.000\\
    0.872 & 0.083 & 0.006 & 0.000\\
    0.873 & 0.222 & 0.004 & 0.153\\
    0.874 & 0.153 & 0.007 & 0.000\\
    0.878 & 0.069 & 0.000 & 0.014\\
    0.879 & 0.264 & 0.067 & 0.264\\
    0.887 & 0.556 & 0.045 & 0.389\\
    0.895 & 0.444 & 0.000 & 0.347\\
    0.900 & 0.458 & 0.042 & 0.306\\
    0.900 & 0.528 & 0.047 & 0.361\\
    0.906 & 0.431 & 0.148 & 0.431\\
    \bottomrule
  \end{tabular}
  \end{threeparttable}
  \end{subtable}%
  \begin{subtable}[t]{.5\linewidth}
  \centering
  \caption{Competitors}
  \label{tab:ozone_rest}
  \begin{threeparttable}
    \begin{tabular}{l r r r}
    \toprule
    \textbf{AUC} & $\mathbf{NF}$  & $\mathbf{NI}$ & $\mathbf{NNM}$ \\
    \midrule
    \multicolumn{4}{c}{\textbf{EBM}}\\
    0.902 & 1.000 & 0.008 & 1.000\\
    \midrule
    \multicolumn{4}{c}{\textbf{Elastic-Net}}\\
    0.894 & 0.792 & 0.000 & 0.000\\
    \midrule
    \multicolumn{4}{c}{\textbf{Random Forest}}\\
    0.839 & 1.000 & 1.000 & 1.000\\
    \midrule
    \multicolumn{4}{c}{\textbf{XGBoost}}\\
    0.915 & 1.000 & 1.000 & 1.000\\
    \bottomrule
    \end{tabular}
  \end{threeparttable}
  \end{subtable}%
\end{table}

\section{Details on the Benchmark Experiments}\label{sec:appendix_benchmark}
We release all code for using \emph{EAGGA} and reproducing our results via \repo.
Benchmark experiments were run on an internal HPC cluster using Intel Xeon E5-2670 instances taking around 17700 CPU hours (benchmarks and ablation study).
Total emissions are estimated to be an equivalent of roughly $1110$ kg CO\textsubscript{2}.
Table~\ref{tab:openml} summarizes all tasks used in our benchmark experiments.
Outer and inner resampling splits were fixed via different random seeds over the 10 replications but the same for all methods.

\begin{table}[ht]
  \centering
  \caption{OpenML tasks used in the benchmarks.}
  \label{tab:openml}
  \footnotesize
  \begin{threeparttable}
  \begin{tabular}{rrrrr}
  \toprule
  & & & \multicolumn{2}{c}{\textbf{Number of}} \\\cline{4-5}
  \textbf{Task ID} & \textbf{Name} & & \textbf{Observations} & \textbf{Features} \\
  \midrule
    37 &                         diabetes & &   768 &   8\\
    43 &                         spambase & &  4601 &  57\\
  3903 &                              pc3 & &  1563 &  37\\
  3904 &                              jm1 & & 10885 &  21\\
  3913 &                              kc2 & &   522 &  21\\
  3918 &                              pc1 & &  1109 &  21\\
  9946 &                             wdbc & &   569 &  30\\
 10093 &          banknote-authentication & &  1372 &   4\\
146819 & climate-model-simulation-crashes & &   540 &  20\\
146820 &                             wilt & &  4839 &   5\\
167120 &                      numerai28.6 & & 96320 &  21\\
168350 &                          phoneme & &  5404 &   5\\
189922 &                             gina & &  3153 & 970\\
190137 &                  ozone-level-8hr & &  2534 &  72\\
190392 &                         madeline & &  3140 & 259\\
190410 &                       philippine & &  5832 & 308\\
359955 & blood-transfusion-service-center & &   748 &   4\\
359962 &                              kc1 & &  2109 &  21\\
359972 &                          sylvine & &  5124 &  20\\
359975 &                        Satellite & &  5100 &  36\\
  \bottomrule
  \end{tabular}
  \begin{tablenotes}
  \tiny
  \item IDs correspond to OpenML task IDs, which enable querying task properties via \url{https://www.openml.org/t/<id>}. Task 3904 originally includes five observations with missing data, which were removed to allow for consistency over all tasks.
  \end{tablenotes}
  \end{threeparttable}
\end{table}

\emph{EAGGA} was configured to use a population size of $\mu = 100$ and an offspring size of $\nu = 10$.
The overall crossover probability was set to $p = 0.7$ and the overall mutation probability to $p = 0.3$.
If crossover was to be applied, each hyperparameter of the search space of the learning algorithm underwent uniform crossover ($p = 0.5$) and crossover of group structures was performed as described in Section~\ref{sec:gga_operators}.
If mutation was to be applied, each numeric and integer hyperparameter of the search space of the learning algorithm underwent Gaussian mutation ($p = 0.2, \sigma = 0.1$; values min-max scaled to $[0, 1]$ prior to mutation and re-transformed afterwards; values rounded to the closest integer in the case of integer hyperparameters), while each categorical hyperparameter underwent uniform mutation ($p = 0.2$), and each group structure was mutated by assigning each feature a new group membership with probability $p = 0.2$ and sampling a new monotonicity attribute for each group with probability $p = 0.2$.
The hyperparameters of the initial population were constructed by using the default hyperparameters of the search space of the learning algorithm, which were then mutated as described above, except for one member of the population which was left unchanged.
The group structures of the initial population were constructed using detectors as described in Section~\ref{sec:init}.
We used the entropy-based information gain feature filter \citep{Largeron_2011} as feature detector, a re-implementation of FAST \citep{lou2012intelligible} using a bin size of $10$ as interaction detector and a monotonicity detector based on Spearman's $\rho$.
Parents were selected via binary tournament selection using non-dominated sorting and crowding distance as criteria.
Parents that resulted in zero features being selected where excluded from the tournament selection.
A $(\mu + \nu)$ survival scheme was used based on non-dominated sorting and crowding distance.

\begin{table}
  \centering
  \caption{EBM search space.}
  \label{tab:ss_ebm}
  \footnotesize
  \begin{threeparttable}
  \begin{tabular}{lrrrr}
  \toprule
  \textbf{Hyperparameter} & \textbf{Type} & \textbf{Range} & \textbf{Trafo} & \textbf{Default} \\
  \midrule
  interactions & int. & $[0, \max (10, \lceil{\sqrt{p(p - 1) / 2}}\rceil)]$ & & 10\\
  outer\_bags & int. & $[8, 50]$ & & 8\\
  inner\_bags & int. & $[0, 50]$ & & 0\\
  max\_rounds & int. & $\{5000, 10000\}$ & & 5000\\
  max\_leaves & int. & $[2, 5]$ & & 3\\
  max\_bins & int. & $[32, 1024]$ & $\log_{2}$ & 256\\
  \bottomrule
  \end{tabular}
  \begin{tablenotes}
  \tiny
  \item ``$\log_{2}$'' in the Trafo column indicates that this parameter is optimized on a (continuous) logarithmic scale with base 2, i.e., the range is given by $[\log_{2}(\mathrm{lower}), \log_{2}(\mathrm{upper})]$, and values are re-transformed to the power of 2 prior to their evaluation. Parameters part of the full EBM search space that are not shown are set to their default. The Default column shows the values recommended by the EBM authors which were always used as the first initial design point.
  \end{tablenotes}
  \end{threeparttable}
\end{table}

\begin{table}
  \centering
  \caption{Elastic-Net search space.}
  \label{tab:ss_glmnet}
  \footnotesize
  \begin{threeparttable}
  \begin{tabular}{lrrr}
  \toprule
  \textbf{Hyperparameter} & \textbf{Type} & \textbf{Range} & \textbf{Trafo}\\
  \midrule
  alpha & cont. & $[0, 1]$ & \\
  s & cont. & $[\exp(-7), \exp(7)]$ & $\log$\\
  \bottomrule
  \end{tabular}
  \begin{tablenotes}
  \tiny
  \item ``$\log$'' in the Trafo column indicates that this parameter is optimized on a (continuous) logarithmic scale, i.e., the range is given by $[\log(\mathrm{lower}), \log(\mathrm{upper})]$, and values are re-transformed via the exponential function prior to their evaluation.
  \end{tablenotes}
  \end{threeparttable}
\end{table}

\begin{table}
  \centering
  \caption{XGBoost search space.}
  \label{tab:ss_xgboost}
  \footnotesize
  \begin{threeparttable}
  \begin{tabular}{lrrrr}
  \toprule
  \textbf{Hyperparameter} & \textbf{Type} & \textbf{Range} & \textbf{Trafo} & \textbf{Default}\\
  \midrule
  nrounds & int. & $[1, 5000]$ & $\log$ & $100$\\
  eta & cont. & $[\num{1e-4}, 1]$ & $\log$ & $0.3$\\ 
  lambda & cont. & $[\num{1e-4}, 1000]$ & $\log$ & $1$\\
  gamma & cont. & $[\num{1e-4}, 7]$ & $\log$ & $\num{1e-4}$\\
  alpha & cont. & $[\num{1e-4}, 1000]$ & $\log$ & $\num{1e-4}$\\
  subsample & cont. & $[0.1, 1]$ & & $1$\\
  max\_depth & int. & $[1, 20]$ & & $6$\\
  min\_child\_weight & cont. & $[1, 150]$ & $\log$ & $\exp(1)$\\
  colsample\_bytree & cont. & $[0.01, 1]$ & & $1$\\
  colsample\_bylevel & cont. & $[0.01, 1]$ & & $1$\\
  \bottomrule
  \end{tabular}
  \begin{tablenotes}
  \tiny
  \item ``$\log$'' in the Trafo column indicates that this parameter is optimized on a (continuous) logarithmic scale, i.e., the range is given by $[\log(\mathrm{lower}), \log(\mathrm{upper})]$, and values are re-transformed via the exponential function prior to their evaluation. Parameters part of the full XGBoost search space that are not shown are set to their default. The Default column shows the initial values used as starting points for the initialization process in \emph{EAGGA}.
  \end{tablenotes}
  \end{threeparttable}
\end{table}

In Section~\ref{sec:benchmarks_competitors}, we compare \emph{EAGGA}\textsubscript{XGBoost} and \emph{EAGGA}\textsubscript{XGBoost\textsubscript{md2}} to an EBM, Elastic-Net, and XGBoost optimized for performance and an untuned random forest.
Search spaces of the learning algorithms are given in Table~\ref{tab:ss_ebm}, Table~\ref{tab:ss_glmnet}, and Table~\ref{tab:ss_xgboost}.
The EBM, Elastic-Net, and XGBoost were optimized via sequential Bayesian Optimization similarly configured as SMAC \citep{lindauer2022smac3}, i.e., using a random forest as surrogate model and Expected Improvement \citep{Jones_Schonlau_Welch_1998} as acquisition function, which was optimized using a random search with a budget of $10000$ function evaluations.
The initial design of size $4d$ ($d$ being the dimensionality of the search space) was sampled uniformly at random -- except for the EBM, where the first initial design point was always given by the default configuration suggested by the EBM authors.
\emph{EAGGA}\textsubscript{XGBoost} and \emph{EAGGA}\textsubscript{XGBoost\textsubscript{md2}} operate on the search space as given in Table~\ref{tab:ss_xgboost}, with the exception that \texttt{max\_depth} was fixed to 2 for \emph{EAGGA}\textsubscript{XGBoost\textsubscript{md2}}.

In some runs, evaluating the initial design during optimization of the EBM took longer than the whole compute budget of 8 hours (mostly for tasks 43, 10093, 189922, 190392, 167120, 190410, 168350, 359972 and 146820).
In these cases, our fallback was to only evaluate the default configuration suggested by the EBM authors.

The random forest was implemented within XGBoost using \texttt{booster = "gbtree"}, \texttt{tree\_method = "exact"}, \texttt{subsample = 1 - exp(-1)}, \texttt{colsample\_bynode = 1 - exp(-1)}, \texttt{num\_parallel\_tree = 1000}, \texttt{nrounds = 1} and \texttt{eta = 1}.

\begin{figure*}[p]
\centering
\includegraphics[width=\textwidth]{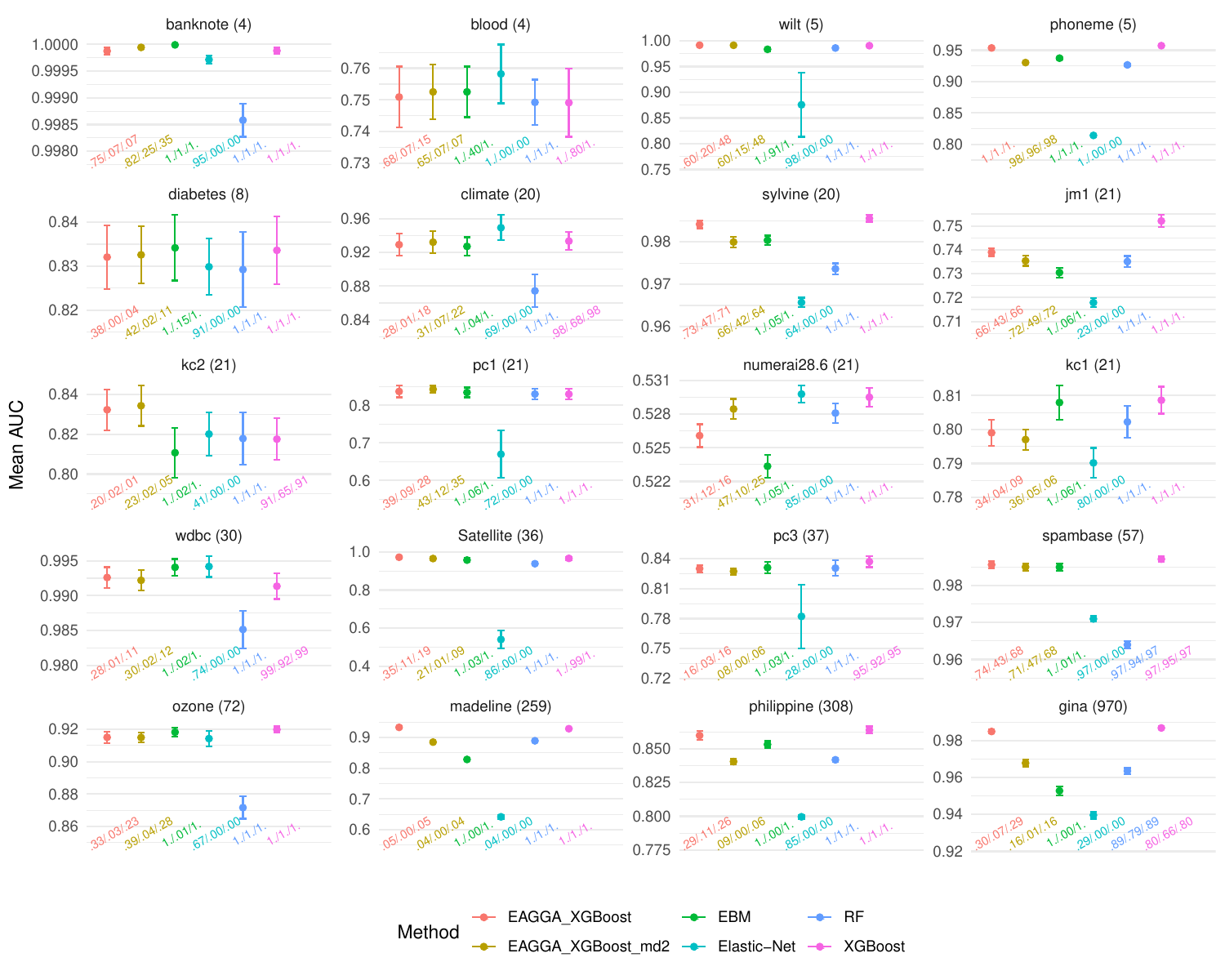}
\caption{Mean AUC of the best models found by each method and their mean interpretability measures ($NF/NI/NNM$ stated below each point) averaged over 10 replications. Bars represent standard errors.}
\label{fig:best_all}
\end{figure*}

In our analysis, we also inspected the best models found by each method and how they perform with respect to the AUC as well as $NF$, $NI$ and $NNM$.
Figure~\ref{fig:best_all} visualizes the mean AUC of these best models.
On the x-axis the average interpretability measures of these models are stated ($NF/NI/NNM$).
We observe that black box models like XGBoost or random forests (RF) often rely on almost all features as well as many interactions of features when solely optimized for performance.
The EBM models often show decent performance using few interactions but all features whereas the Elastic-Net models can be very sparse but often lack good performance. 
In contrast, the best models found by \emph{EAGGA}\textsubscript{XGBoost} and \emph{EAGGA}\textsubscript{XGBoost\textsubscript{md2}} often perform almost on par with the XGBoost models solely optimized for performance but use substantially fewer features and interactions and often also rely on fewer non-monotone features.
For example, on the \emph{philippine} task, the best models found by \emph{EAGGA}\textsubscript{XGBoost} result in an average AUC of around $0.860$ using on average $29\%$ of features, $11\%$ of feature interactions and on average only $26\%$ of the features have a non-monotone effect.
In contrast, the XGBoost models optimized for performance result in an average AUC of around $0.864$ but on average use all features, include interactions of all features and cannot guarantee that some of the features used in the model are restricted to have a monotone effect.

XGBoost\textsubscript{MO} in Section~\ref{sec:benchmarks_xgboost_mo} was optimized via ParEGO using a random forest as surrogate model and Expected Improvement as acquisition function, which was optimized using a random search with a budget of $10000$ function evaluations.
The search space is given in Table~\ref{tab:ss_xgboost}.
The initial design of size $4d$ was sampled uniformly at random.

\begin{figure*}[p]
\centering
\includegraphics[width=\textwidth]{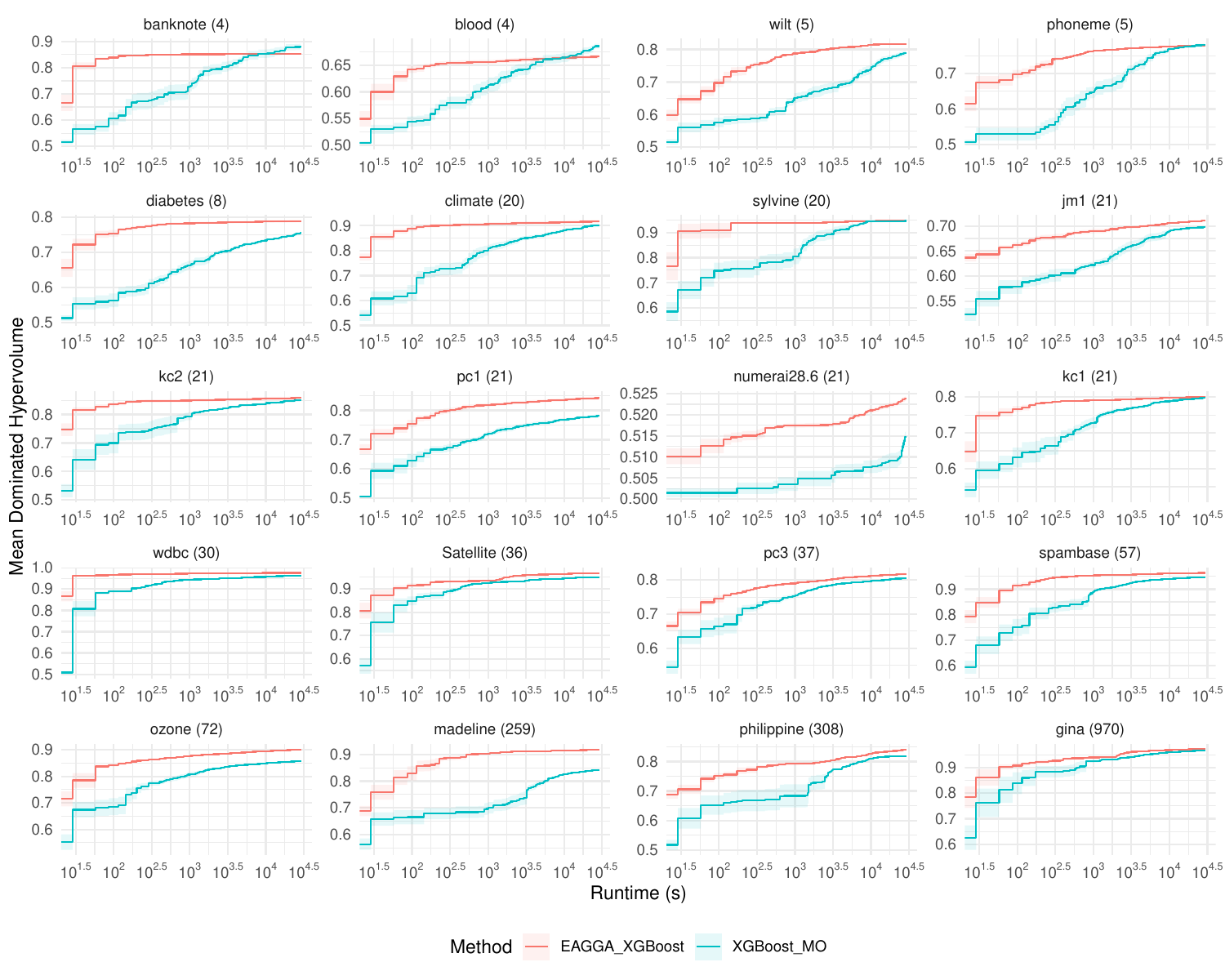}
\caption{Anytime mean dominated Hypervolume during optimization of \emph{EAGGA}\textsubscript{XGBoost} and XGBoost\textsubscript{MO} averaged over 10 replications. Ribbons represent standard errors. Note that the x-axis is shown on $\log_{10}$ scale.}
\label{fig:eagga_ablation_xgboost}
\end{figure*}

We also computed the anytime dominated Hypervolume during optimization (i.e., calculated on the inner resampling), see Figure~\ref{fig:eagga_ablation_xgboost}.
Notably, the only tasks where XGBoost\textsubscript{MO} eventually outperforms \emph{EAGGA}\textsubscript{XGBoost} are low-dimensional tasks with four or five features.
A Wilcoxon signed-ranks test on the final mean dominated Hypervolume indicates that \emph{EAGGA}\textsubscript{XGBoost} indeed solves the inner optimization problem better than XGBoost\textsubscript{MO} ($T = 30, p = 0.0026$).

\section{Details on the Ablation Study of \emph{EAGGA}}\label{sec:appendix_ablation}

\begin{figure*}[p]
\centering
\includegraphics[width=\textwidth]{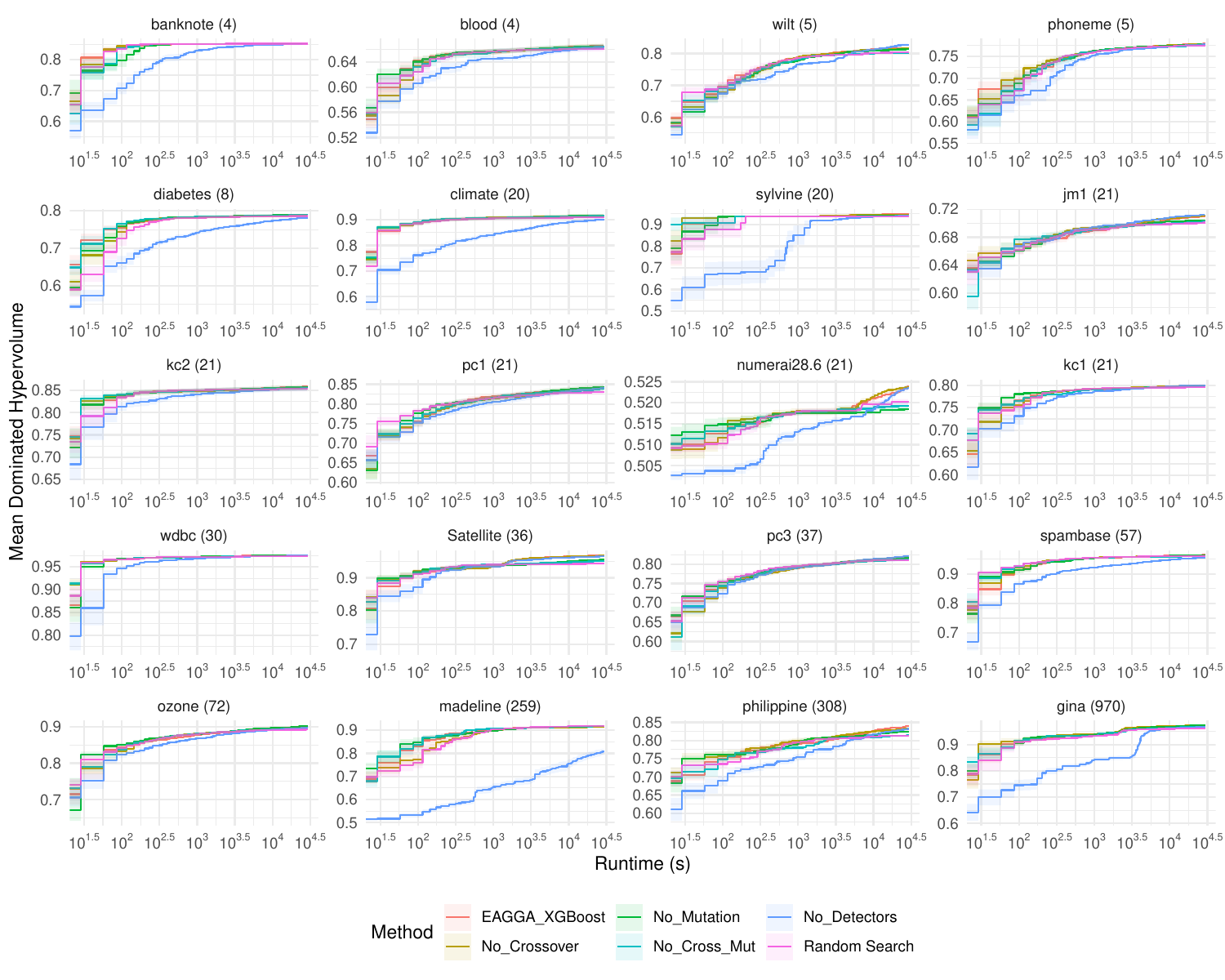}
\caption{Anytime mean dominated Hypervolume during optimization of \emph{EAGGA}\textsubscript{XGBoost} and different flavors averaged over 10 replications. Ribbons represent standard errors. Note that the x-axis is shown on $\log_{10}$ scale.}
\label{fig:eagga_ablation}
\end{figure*}

Here, we report detailed results of the ablation study of \emph{EAGGA}.
Figure~\ref{fig:eagga_ablation} visualizes the anytime mean dominated Hypervolume during optimization (i.e., calculated on the inner resampling) of \emph{EAGGA}\textsubscript{XGBoost} and different flavors as compared in our ablation study.
Table~\ref{tab:eagga_ablation_final} shows the corresponding final mean dominated Hypervolume.
We observe that using detectors often results in a strong performance boost, but \emph{EAGGA} without detectors usually catches up in performance.
Generally, not performing either crossover or mutation of group structures results in comparably poor performance, and performing neither crossover nor mutation results in final performance close to the random search.

We hypothesize that the effectiveness of using or not using crossover or mutation may depend on the performance of the detectors used to initialize the population in \emph{EAGGA}.
If the initial group structures determined by the detectors are already high-performing, then crossover during optimization could hinder progress due to excessive exploration.
In such cases, using only mutation of the initial group structures may be more effective.
On the other hand, if detector performance is poor, more exploration of group structures may be needed, and crossover during optimization can be beneficial.
In conclusion, we believe that the performance of \emph{EAGGA} can be further significantly enhanced by fine-tuning the configuration of mutation and crossover rates.
Additionally, considering the possibility of changing these rates in a self-adaptive manner \citep{Li_2013} could further improve the optimization process.

\begin{table*}[p]
\centering
  \caption{Final mean dominated Hypervolume of \emph{EAGGA}\textsubscript{XGBoost}, different flavors and XGBoost\_MO during optimization.}
  \label{tab:eagga_ablation_final}
  \footnotesize
  \begin{threeparttable}
  \begin{tabular}{rrrrrrrrr}
  \toprule
  \textbf{Task ($p$)} & & \multicolumn{7}{c}{\textbf{Method}} \\ \cmidrule(r){3-9}
  & & EAGGA\_XGBoost & No\_Crossover & No\_Mutation & No\_Cross\_Mut & No\_Detectors & Random Search & XGBoost\_MO\\ 
  \midrule
  banknote (4)     & & 0.853 (0.000) & 0.852 (0.000) & 0.852 (0.000) & 0.853 (0.000) & 0.852 (0.001) & 0.852 (0.001) & \textbf{0.880} (0.012) \\ 
  blood (4)        & & 0.666 (0.004) & 0.665 (0.004) & 0.665 (0.004) & 0.664 (0.004) & 0.664 (0.004) & 0.661 (0.004) & \textbf{0.686} (0.005) \\ 
  wilt (5)         & & 0.817 (0.001) & 0.816 (0.002) & 0.814 (0.003) & 0.803 (0.003) & \textbf{0.828} (0.004) & 0.805 (0.001) & 0.789 (0.007) \\ 
  phoneme (5)      & & 0.779 (0.001) & 0.778 (0.001) & 0.777 (0.001) & 0.775 (0.001) & 0.777 (0.001) & 0.775 (0.001) & \textbf{0.779} (0.004) \\ 
  diabetes (8)     & & 0.789 (0.003) & 0.788 (0.003) & \textbf{0.789} (0.003) & 0.789 (0.003) & 0.782 (0.003) & 0.786 (0.003) & 0.757 (0.005) \\ 
  climate (20)     & & 0.917 (0.006) & 0.916 (0.006) & \textbf{0.917} (0.006) & 0.916 (0.006) & 0.902 (0.008) & 0.910 (0.007) & 0.902 (0.007) \\ 
  sylvine (20)     & & \textbf{0.949} (0.001) & 0.947 (0.001) & 0.940 (0.001) & 0.940 (0.001) & 0.941 (0.003) & 0.939 (0.001) & 0.946 (0.004) \\ 
  jm1 (21)         & & 0.710 (0.002) & 0.710 (0.001) & 0.704 (0.002) & 0.702 (0.002) & \textbf{0.712} (0.001) & 0.701 (0.002) & 0.697 (0.002) \\ 
  kc2 (21)         & & \textbf{0.858} (0.005) & 0.855 (0.005) & 0.857 (0.004) & 0.855 (0.005) & 0.855 (0.005) & 0.853 (0.005) & 0.851 (0.005) \\ 
  pc1 (21)         & & 0.843 (0.006) & 0.839 (0.006) & \textbf{0.843} (0.007) & 0.838 (0.006) & 0.840 (0.006) & 0.830 (0.005) & 0.782 (0.006) \\ 
  numerai28.6 (21) & & \textbf{0.524} (0.001) & 0.524 (0.000) & 0.518 (0.001) & 0.519 (0.001) & 0.524 (0.001) & 0.520 (0.001) & 0.515 (0.001) \\ 
  kc1 (21)         & & 0.800 (0.002) & 0.800 (0.002) & 0.798 (0.002) & 0.797 (0.002) & \textbf{0.800} (0.002) & 0.796 (0.002) & 0.797 (0.003) \\ 
  wdbc (30)        & & \textbf{0.976} (0.001) & 0.975 (0.001) & 0.976 (0.000) & 0.974 (0.001) & 0.975 (0.001) & 0.974 (0.001) & 0.964 (0.003) \\ 
  Satellite (36)   & & \textbf{0.968} (0.002) & 0.967 (0.002) & 0.955 (0.003) & 0.951 (0.003) & 0.965 (0.002) & 0.944 (0.004) & 0.950 (0.004) \\ 
  pc3 (37)         & & 0.818 (0.003) & 0.816 (0.004) & 0.817 (0.003) & 0.812 (0.003) & \textbf{0.821} (0.004) & 0.811 (0.004) & 0.805 (0.004) \\ 
  spambase (57)    & & \textbf{0.964} (0.001) & 0.962 (0.000) & 0.964 (0.001) & 0.960 (0.000) & 0.954 (0.002) & 0.961 (0.001) & 0.947 (0.001) \\ 
  ozone (72)       & & 0.901 (0.002) & 0.899 (0.002) & \textbf{0.901} (0.002) & 0.895 (0.002) & 0.897 (0.002) & 0.892 (0.003) & 0.858 (0.003) \\ 
  madeline (259)   & & 0.918 (0.002) & 0.915 (0.002) & 0.918 (0.003) & 0.918 (0.002) & 0.807 (0.013) & \textbf{0.919} (0.002) & 0.842 (0.004) \\ 
  philippine (308) & & \textbf{0.839} (0.003) & 0.832 (0.003) & 0.825 (0.003) & 0.814 (0.002) & 0.832 (0.003) & 0.812 (0.002) & 0.818 (0.001) \\ 
  gina (970)       & & 0.972 (0.001) & \textbf{0.972} (0.001) & 0.972 (0.001) & 0.966 (0.001) & 0.967 (0.001) & 0.962 (0.002) & 0.967 (0.001) \\ 
  \bottomrule
  \end{tabular}
  \begin{tablenotes}
  \tiny
  \item Best final mean dominated Hypervolume highlighted in bold. Standard error over 10 replications in parentheses. "(0.000)" denotes that the standard error is smaller than 0.0005. For completeness we also include XGBoost\_MO.
  \end{tablenotes}
  \end{threeparttable}
\end{table*}

\end{document}